\documentclass[a4paper,fleqn]{cas-dc}

\usepackage[numbers,sort&compress]{natbib}

\def\tsc#1{\csdef{#1}{\textsc{\lowercase{#1}}\xspace}}
\tsc{WGM}
\tsc{QE}

\begin{document}
\let\WriteBookmarks\relax
\def\floatpagepagefraction{1}
\def\textpagefraction{.001}

\shorttitle{Novel End-to-End Production-Ready Machine Learning Flow for Nanolithography Modeling and Correction}    

\shortauthors{Mohamed Habib, Hossam Fahmy, Mohamed Abu-Elyazeed}  

\title [mode = title]{Novel End-to-End Production-Ready Machine Learning Flow for Nanolithography Modeling and Correction}


%

\author[1]{Mohamed S. E. Habib}[type=editor,
	   role=Researcher,
       style=English,
       auid=000,
       bioid=1,
	   prefix=]

\ead{mohamed1611071@eng1.cu.edu.eg}



\affiliation[1]{organization={EECE, Faculty of Engineering, Cairo University},
            country={Egypt}}

\author[1]{Hossam A. H. Fahmy}[role=Researcher,
       auid=000,
       bioid=1,
       prefix=]

\ead{hossam.fahmy@eng.cu.edu.eg}

\author[1]{Mohamed F. Abu-ElYazeed}[role=Researcher,
       auid=000,
       bioid=1,
       prefix=]
       

\ead{mfathyae@eng.cu.edu.eg}







\begin{abstract}
Optical lithography is the main enabler to semiconductor manufacturing. It requires extensive processing to perform the Resolution Enhancement Techniques (RETs) required to transfer the design data to a working Integrated Circuits (ICs). The processing power and computational runtime for RETs tasks is ever increasing due to the continuous reduction of the feature size and the expansion of the chip area. State-of-the-art research sought Machine Learning (ML) technologies to reduce runtime and computational power, however they are still not used in production yet. In this study, we analyze the reasons holding back ML computational lithography from being production ready and present a novel highly scalable end-to-end flow that enables production ready ML-RET correction.
\end{abstract}




\maketitle





\section{Introduction}\label{intro}
Computational lithography aims to prepare synthesized integrated circuits (ICs) designs for accurate transfer into the semiconductor wafer. It performs elaborate correction flows that adds, subtracts and modifies the original IC patterns with the end goal of minimizing the fabrication process errors. Such correction flows and operations are often referred to as Resolution Enhancement Techniques (RETs) and are known for being computationally expensive and time consuming.
 
As nanolithography technologies continue to advance towards ever decreasing minimum feature sizes, the same die area is able to host more design data. Henceforth, the data volume required to produce the same wafer area increase substantially, consequently, increasing the computational power and runtime required to perform the RET correction.

Computational lithography industry sought Machine Learning (ML)\footnote{We use the term ``machine learning'' as generalized term to refer to any advanced AI techniques including deep learning, generative networks and similar technologies.}
algorithms to accelerate the RET correction and reduce the computational load. However, ML-RET are not yet adopted for production to the best knowledge of the authors.

In this study, we analyze the reasons holding back ML-RET correction from utilization in RET production flows. Based on this analysis, we introduce a novel flow that mitigates those obstructions and offers an end-to-end production-ready platform for ML-RET with very high scalability. The rest of this paper is organized as follows: in section \ref{trad}, we discuss the traditional way of performing RET correction. Next, we discuss and analyze the state-of-the-art of ML-RET and the roadblocks holding it from being a viable production option in section \ref{probs}. We then introduce TPM-RET, a novel end-to-end production friendly ML flow to model and correct lithography in section \ref{TPMRET}, and showcase some of its results in section \ref{cs}. Next, we discuss how TPM-RET flow solves the production difficulties in section \ref{discuss}. Finally, we present our future plans and conclusions in sections \ref{future} and \ref{conclusions} respectively.

\section{Traditional RET}\label{trad}
Optical lithography machinery provided good accuracy early-on for large pattern dimensions. However, its accuracy and fidelity degraded as manufacturing requirements progressed towards smaller dimensions, even though such dimensions are still within the process theoretical limits. Literature indicates\cite{mackFundamentalPrinciplesOptical2007} that the influence of Optical, mechanical and chemical effects become more impactful as the printed features progress towards the theoretical lithography limits.

It is impractical to replace the production lines for every new lithography node, given the hefty time and monetary investments put into research and yield stabilization. Thus, it is the duty of advanced computational lithography and RET techniques to extend the lifespan of the existing infrastructure with little or no modification to the hardware.

RET techniques were invented gradually one after another as the need arose and extra correction steps were required. Such techniques are executed sequentially in standard tapeout flows that grew more complex with every technology advancement. Furthermore, each one of these techniques grew more complicated as the accuracy degradation grew more prominent. Here is a list of some famous RET techniques\cite{wongResolutionEnhancementTechniques2001,linModelBasedMask2009}:
\begin{itemize}
	\item Optical Proximity Correction (OPC)
	\item Sub-Resolution Assist Features (SRAFs)
	\item Off-Axis Illumination (OAI)
	\item Mask Process Correction (MPC)
	\item Phase-Shifing Masks (PSM)
\end{itemize}

In general, a RET technique operates in one of the following ways:
\begin{enumerate}
	\item Modify the mask shape by moving its edges or introduce new features such that the results of the mask illumination is as close as possible to original target shapes.
	\item Change the laser source shape or incidence angle to enhance contrast at wafer plane or give optimal illumination conditions to dominant mask feature configuration.
	\item Alter photomask stack in order to introduce phase difference between bright and dark areas consequently offering better contrast for the wafer image.
\end{enumerate}

The ever-growing computational overhead of traditional RET drive the industry to look for a faster and less computationally demanding approaches. ML-RETs offer an attractive alternative to reduce the computational cost of the RETs, however such solutions are not yet adopted for RET production flows.

\section{ML-RET State-of-the-Art Analysis}\label{probs}
In this section, we briefly discuss the current state-of-the-art ML-RET solutions and analyze the obstacles holding back this technology from reaching production-level despite the efforts expended.

\subsection{State-of-the-Art Synopsis}
Literature shows clear research directions and preferred methods for performing ML-RET correction. These directions are influenced by the nature of the RET correction itself as well as the latest advancements in ML industry especially in the image processing field. Based on our research, we summarize the key characteristics of the ML-RET meta as follows:

\subsubsection{Image-Based Photomask Correction}
Recent state-of-the-art ML-RET methods use image-based input by converting design patterns into image slices\cite{yangGANOPCMaskOptimization2018,chenDevelSetDeepNeural2021,jiangNeuralILTMigratingILT2022,ciouMachineLearningOPC2022,chenDAMODeepAgile2022,alawiehGANSRAFSubresolutionAssist2021}. The ML model then processes such input and transfers it to the corrected photomask domain to output the final photomask images. 

\subsubsection{Generative ML Techniques}
Generative Adversarial Networks (GANs)\cite{goodfellowGenerativeAdversarialNetworks2014} and other advanced domain transfer ML techniques are the mainstream methods for state-of-the-art ML-RET research\cite{yangGANOPCMaskOptimization2018,chenDevelSetDeepNeural2021,jiangNeuralILTMigratingILT2022,ciouMachineLearningOPC2022,chenDAMODeepAgile2022,alawiehGANSRAFSubresolutionAssist2021,ciouSRAFPlacementGenerative2021}. This can be attributed to the great advancements of image generation and translation techniques in recent years, henceforth, encouraging their usage to solve ML-RET as a domain transfer problem.

\subsubsection{Single RET ML Applications}
State-of-the-Art research focuses on ML specific treatment targeting a single RET application\cite{yangGANOPCMaskOptimization2018,chenDevelSetDeepNeural2021,jiangNeuralILTMigratingILT2022,ciouMachineLearningOPC2022,chenDAMODeepAgile2022,alawiehGANSRAFSubresolutionAssist2021,ciouSRAFPlacementGenerative2021,matsunawaOpticalProximityCorrection2016,kwonOpticalProximityCorrection2019,gengSRAFInsertionSupervised2020,shimEtchProximityCorrection2016,mengMachineLearningModels2021,sharmaMachineLearningGuided2019}. OPC correction and SRAF insertion are the most researched techniques in that regards, since they are the most challenging and computationally expensive operations in the production tapeout flow.

\subsubsection{Model-Based Reference Data}
Reference data required for ML model training is usually generated using state-of-the-art model-based RET solutions\cite{yangGANOPCMaskOptimization2018,chenDevelSetDeepNeural2021,jiangNeuralILTMigratingILT2022,ciouMachineLearningOPC2022,chenDAMODeepAgile2022,alawiehGANSRAFSubresolutionAssist2021,matsunawaOpticalProximityCorrection2016,kwonOpticalProximityCorrection2019,gengSRAFInsertionSupervised2020,shimEtchProximityCorrection2016,mengMachineLearningModels2021,sharmaMachineLearningGuided2019}. Inverse Lithography Technology (ILT)\cite{pangInverseLithographyTechnology2021} is often used to fulfill this role as it provides the theoretical best photomask for a given pattern\cite{maFastPixelbasedOptical2014,yangGANOPCMaskOptimization2018,chenDevelSetDeepNeural2021,jiangNeuralILTMigratingILT2022}.

\subsection{Difficulties Facing Production-Ready ML-RET}\label{diff}
\subsubsection{Loss of Mask Information}
The final photomask after RET correction is a binary pattern, where any pixel on it can be either bright or dark. Traditionally, the final photomask is obtained through a series of simulations and iterative correction until reaching acceptable accuracy.

For ML-RET model training, the binary photomask data is used to teach the model about the correction process and ideal corrected patterns. However, this means losing valuable information about the intermediate operations and obscures a lot of the physical and process relations. This, in turn, makes it hard for the ML-RET model to infer the correct relations governing the process behavior. Thus, the risk of over training increases and the reduces the ML models ability to handle never-seen patterns. 

\subsubsection{Full-Chip Scale Issues}
State-of-the-Art research focuses on the accuracy and runtime aspects of calibration and evaluation of ML-RET models. To the best knowledge of the authors, not much attention goes to enabling chip scale evaluation, which is a key aspect for enabling the technology. Since a full chip is too large to be processed as a single image, the need arise for splitting the chip into smaller slices and stitching them back after correction is done. 

For a ML-enabled process, slicing the chip data is not a simple task especially for when using GANs and image translation techniques. The recurring patterns should be put consistently in the same location with respect to its window slice to make sure the correction results are consistent.

The reassembly of the chip from the corrected images slices is yet another nontrivial task that calls for complex post-processing rules. First, it requires stitching back polygons spanning across multiple windows together, and consequently, resolving any conflicts that appear at window borders. Furthermore, more post processing is required to resolve any MRC\footnote{Mask Rule Checks: Special rules defined by mask house to make sure the photomask is manufacturable.} violations that may appear if polygons in separate windows come too close after stitching them together.

\subsubsection{Correction Consistency}\label{consistency}
Consistency is an important aspect for IC manufacturing to ensure similar treatment for same patterns across a chip scale correction. This is especially desired for designs based on symmetrical device topologies in order to ensure uniform device performance, reliability and minimize electrical stress.

The topological features around the target pattern is a main factor that dictates its correction. Slight shifts in the correction window, i.e due to chip slicing in an GAN-based correction, can lead to a different ML-RET response to the same pattern resulting into inconsistent correction. 

The stitching operations during full-chip assembly can be another source of correction inconsistency. This happens in the cases where the post-processing algorithm applies different treatments to the polygons near windows border while resolving stitching conflicts or MRC violations.

\subsubsection{Hardware Requirements}
Traditional RET correction requires massive computational resources, hence FABs usually own huge processing grids with thousands of CPU cores as well as advanced monitoring and allocation management systems. ML models, however, are best trained and utilized in a GPU-based environment.

To adopt state-of-the-art ML-RET flows, FABs need to replace their CPU-based infrastructure and reinvest in a GPU-based one. This creates great deal of resistance in the adoption of ML-RET due to the accumulated experience, time and fund investments in the older infrastructure.

\subsubsection{Pattern Interaction Distance}
The physical phenomena affecting the pattern transfer from photomask to wafer depends heavily on the interactions between neighboring patterns. Computational lithography tools usually have a specified distance where such pattern interactions have an impactful contribution, usually referred to as interaction distance or radius.

Traditional RET tools calculate the interaction distance uniformly around each pattern. This is also true for ML-RET models that utilize uniformly collected features around the corrected pattern, such as models based on Support Vector Machines (SVM), as the feature collection algorithm can be programmed to restrict measurements to the correct interaction distance.

On the other hand, GANs and similar ML-RET models correct for a complete windows as whole, thus the whole window can be considered as interacting features to any pattern inside it. This poses two issues, first, patterns at window edges will have unsymmetrical interaction distance. The second issue is that similar patterns will have variable interaction distance according to their position in the window. This can lead to variability in the correction results and further contribute to inconsistency.

\subsubsection{Input resolution}
Traditional RET techniques as well as feature-based ML-RET, use the IC design data directly to perform the correction or measure layout features respectively. The layout resolution in this case determines the refinement of the edge displacement which can bring forth better accuracy.

GAN-based ML-RET models, contrarily, require the conversion of the IC layout to image to be able to process the data. This adds an extra dimension to the correction process. Using too coarse image resolution sabotages the accuracy of the correction due to the information loss, making the model unable to differentiate between pattern configuration. While using too fine image resolution harms the correction runtime as well as increases the chance of model overfitting and losing the ability to generalize over the input samples.

\subsubsection{Never-Seen Pattern Integration}
Traditionally, a never-seen pattern does not impact process modeling. If such pattern failed to post RET verification, the RET correction recipe is then fine tuned to account for it, or in some rare occasions, it can be entirely disallowed using design rules and hotspot detection tools.

Ml-RET models, on the other hand, are built based on the model experience with patterns that are relatively similar to real-life designs. Hence, model retraining can be necessary to account for such patterns. RET models are the most central part of the correction process, their calibration and fine tuning takes effort, experience and most importantly test photomasks to perfect. Therefore, having to retrain the models to include patterns that are outside of the seen data space is a huge overhead.

\subsubsection{Partitioning of RET Correction}\label{part}
Traditionally, RET correction is done in stages in reverse order of the process manufacturing steps. This staged approach is due to RET techniques being introduced overtime and the CAD industry formulating the correction using such techniques as separate building blocks comprising a big flow.

This staged approach is, however, heavily influencing the correction flow in ML-RET state-of-the-art research. The partition of the ML-RET flow adds extra multiple layers of complexity, requiring dataset preparation, data engineering, model training and verification for every part of the correction flow.

\section{TPM-RET: Novel Production-Friendly ML-RET Flow}\label{TPMRET}
We present True Pixel-based Machine-learning RET (TPM-RET), a novel production-friendly flow for the modeling and correction of the lithography process using convolutional neural network. TPM-RET framework addresses the concerns discussed in section \ref{probs}, providing a CPU-scalable, end-to-end, consistent solution all while being full-chip ready.

In the next subsections, we discuss the design choices defining the TPM-RET flow. The main focus of such discussions is to demonstrate the benefits of this novel flow and how it addresses the aspects that makes ML-RET not appealing for mass-production.

\subsection{Design Choice \#1: True Pixel-Based Correction}\label{TPM}
Modern state-of-the-art methods use GANs to correct chunks of the photomask at one go using input image clips of the design data. This correction style is, in fact, pixel-based, however it generates conflicts at split boundary and consistency distrust due to correction window shifts as previously discussed in section \ref{consistency}. To avoid such issues, gain more control over the correction scheme and allow for flexible image resolution, we opt to apply correction for each pixel of the photomask separately, one at a time.

\subsection{Design Choice \#2: Minimum Model Footprint}\label{MMF}
To reduce any runtime issue from the pixel-by-pixel correction, we choose to use the minimum model that yields good accuracy. Thus, we choose Convolutional Neural Network (CNN) as the model structure of choice for TPM-RET, due to its efficiency in image classification and low computational power requirements.

\subsection{Design Choice \#3: Inverse Intensity Profile (IIP)}\label{IIP}

\begin{figure*}[htp]
	\centerline{\includegraphics[width=0.70\textwidth]{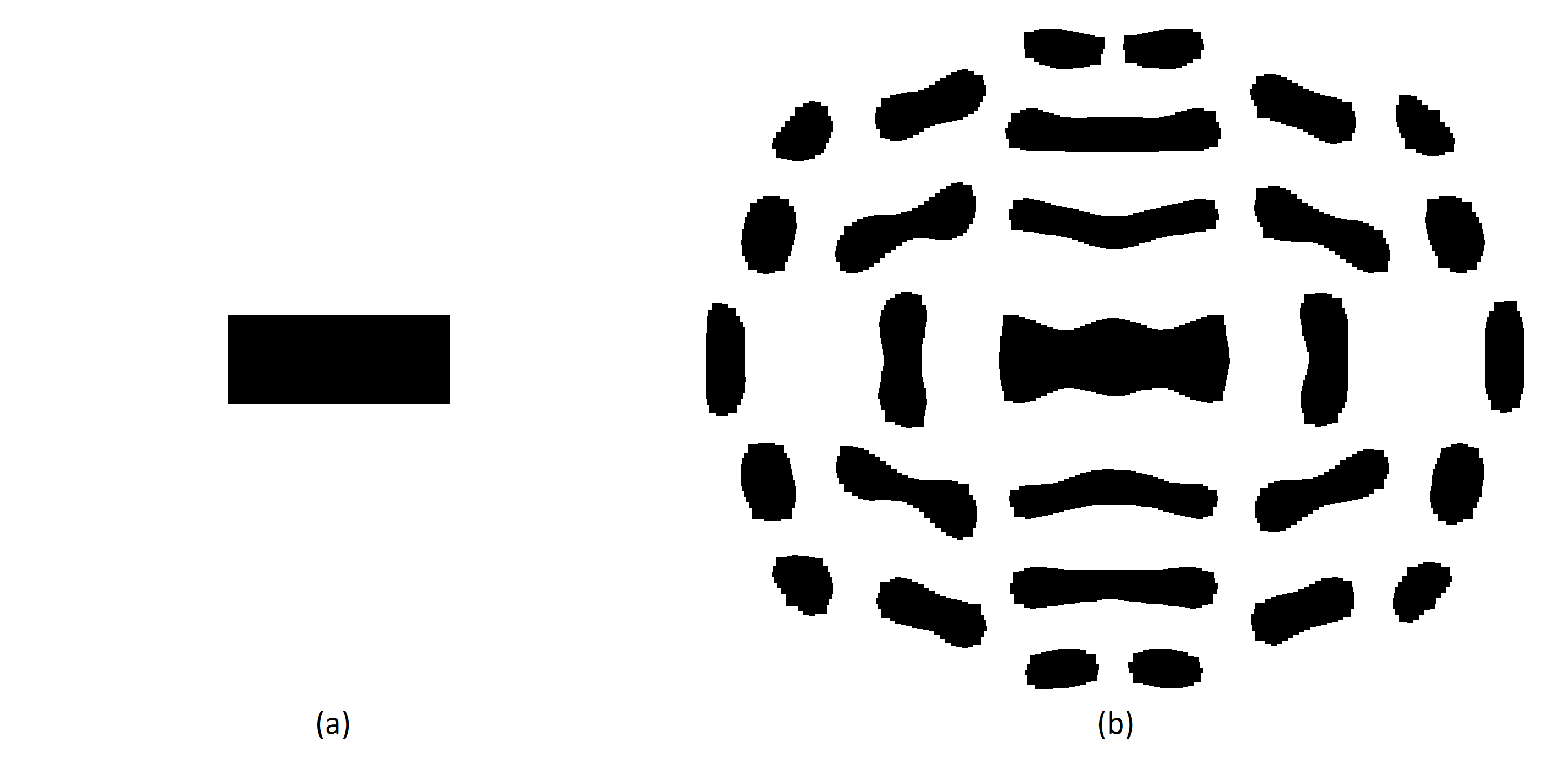}}
	\caption{Example of a simple target wafer pattern \(T\) (a), and its corresponding inverse lithography ideal photomask \(M^*\) (b)}
	\label{ilt}
\end{figure*}

Inverse Lithography Technology [ILT]\cite{pangInverseLithographyTechnology2021} solves the lithography optimization problem inversely. Assuming that we have the ideal silicon pattern \(T(x,y)\) at the wafer level, we calculate the optimum photomask pattern \(M^*(x,y)\) that when put at the scanner will yield such ideal silicon pattern.
In other words, the ILT tries to solve the lithography problem by calculating the inverse of the manufacturing process transfer function \(F()\). Hence, we can calculate the optimal photomask \(M^*\) by solving the equation \eqref{eq2}:
\begin{equation}
	M^*(x,y) = F^{-1}(T(x,y)) 
	\label{eq2}
\end{equation}

Fig. \ref{ilt} shows an example test pattern and its corresponding ILT photomask calculated using Calibre\textsuperscript{\textregistered} pxOPC\textsuperscript{\textregistered}\cite{siemensedaCalibreComputationalLithography2023}. While the \(M^*(x,y)\) is the clearest option for photomask pixel classification, the binary nature of the photomask conceals the process related information which makes it challenging for the model to infer it while training.

In order to recover some of the process information from that binary photomask function, we define Inverse Intensity Profile [IIP] as the spacial convolution product of the binary mask function \(M^*(x,y)\) and a predefined Inverse Intensity Kernel (IIK) \(K(x,y)\) as shown in \eqref{eq1}:

\begin{equation}
	IIP(x,y) = \frac{M^*(x,y) \circledast K(x,y)}{||M^*(x,y) \circledast K(x,y)||}
	\label{eq1}
\end{equation}

\begin{figure}[htp]
	\centerline{\includegraphics[width=0.45\textwidth]{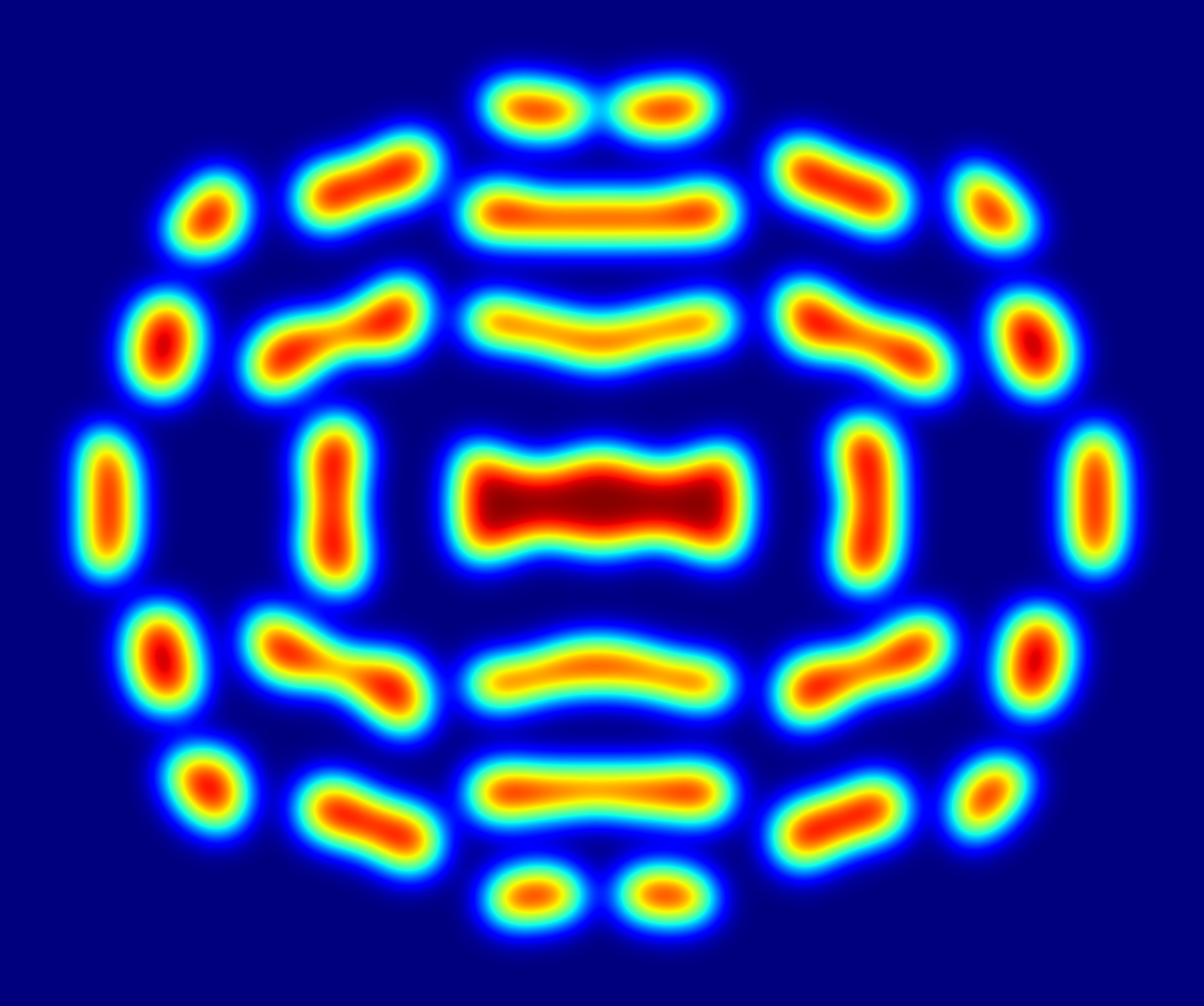}}
	\caption{IIP map corresponding to target pattern in fig. \ref{ilt}.}
	\label{rip}
\end{figure}

The function \(IIP(x,y)\) is a continuous function within the \([0,1]\) range and it provides information about the photomask transitions between ''opaque'' and ''transparent'' states. The kernel \(K(x,y)\) is an optimization parameter that represents the process characteristics. Fig. \ref{rip} shows the \(IIP\) map corresponding to the pattern in shown in fig. \ref{ilt}. Using the introduced \(IIP\) allows the model to learn the process underlying physics during the training phase and hence have more stability and coherence than when using the original binary function.

\subsection{Design Choice \#4: Nonuniform Image Compression}\label{NIC}
A halo region around corrected pattern, a single pixel in our case, is required to provide spacial awareness about its neighborhood during the correction process. Usually, such halo is referred to as Interaction Distance (ID) and it is the furthest distance at which another pattern can affect the correction process. ID is an optimization parameter and is selected based on the required level of accuracy in the final photomask. For a standard 32nm immersion lithography process, typical ID is around 1.2um and can be tuned according to accuracy requirements.

Assuming we take a resolution of 1 pixel per \(1nm^2\), an image of size \(2001x2001\) pixels is required to account for a ID of \(1um\) for every processed pixel. Such image size is too large to use as input for a ML model as it can slow the runtime and cause overfitting. Increasing the number of pixels per \(nm^2\) will further distend the problem.

To fix this issue, we use nonuniform image compression that uses a different compression algorithm for vertical and horizontal pixel arrangements. This is to retain any directional asymmetries in the lithography process and allow the model to distinguish different pattern orientations more easly. Using this compression technique, an image of size \(2001x2001\) will be reduced to \(250x250\) pixels.

\subsection{Design Choice \#5: End-to-End Correction}\label{E2E}
Partitioning the ML-RET correction flow adds unnecessary complexity while lacking the historical justifications that made such approach acceptable for traditional RET flows. Furthermore, cascading the output of ML-RET into another can magnify the error. We opt to executing the whole ML lithography correction in one shot as an end-to-end flow, especially that the required input data for all ML-RET stages already exist in the input images and the IIP map. The benefits of using an end-to-end flow also includes the simplification of dataset engineering, reduction of training and verification times, lenient hardware requirements, and appeals to the ML models power to fit complex patterns.

\subsection{TPM-RET Flow Assembly}\label{FA}
Advancing from the previous discussion, we assemble the design pieces into the TPM-RET flow. We can split the flow into two phases:

\begin{figure*}[bhtp]
	\centerline{\includegraphics[width=0.75\textwidth]{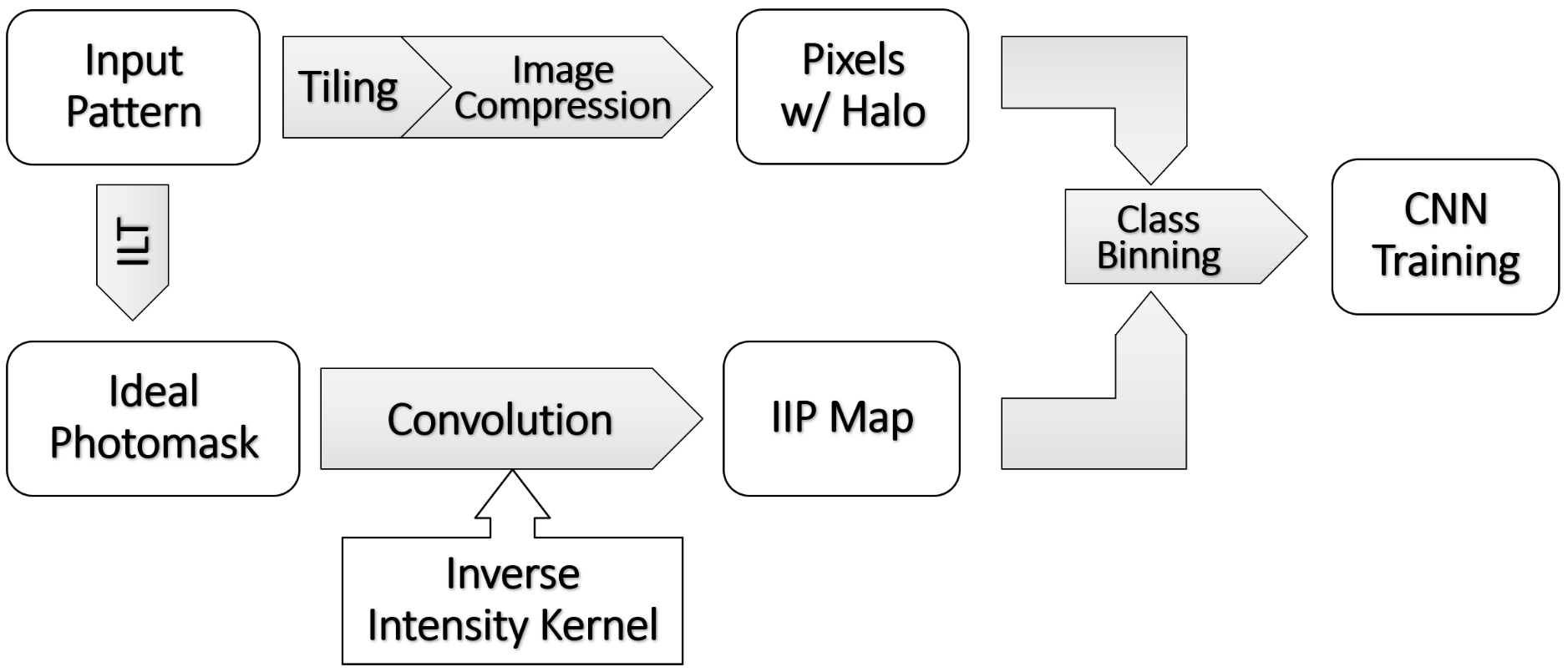}}
	\caption{Flow chart for training data preparation.}
	\label{train}
\end{figure*}

\subsubsection{Data Preparation and Model Training}\label{DPT}
Fig.\ref{train} shows the flow chart for the data preparation phase for the TPM-RET platform. The input pattern is used to extract the pixel images by first converting it into a high resolution image then tile and compress it in order to obtain the images corresponding to the training pixels data. 

Simultaneously, the test pattern is corrected via Calibre\textsuperscript{\textregistered} pxOPC\textsuperscript{\textregistered} to obtain the ILT photomask. The IIP map is obtained as a function of the photomask domain, \(IIP(x,y)\), by convolving the selected IIK \(K(x,y)\) with the ILT photomask \(M^*(x,y)\). 
 
The images are then paired to their corresponding IIP value using their pixel coordinates, such that an image corresponding to pixel \(p(x_p,y_p)\) is coupled with the value \(IIP(x_p,y_p)\). The obtained data are next assigned to the classes according to their \(IIP(x,y)\) value and saved for ML model training. For the sake of this study, we use 100 IIP classes. The number of classes depends on whether the intended application requires a fine IIP map or not. Increasing the number of classes does not affect the runtime for same samples count, but will also require more training samples to populate all the classes. 

Once the data collection is finished, we use the Tensorflow \cite{tensorflow2015} and Keras \cite{chollet2015keras} libraries to construct and train the CNN model. The training data are split into three unique sets, the training and validation sets are used for the model training loop while the third set is used for testing the calibrated model and ensure its accuracy after the training is completed.

\subsubsection{Model Deployment and Post Processing}\label{RPP}
\begin{figure*}[bhtp]
	\centerline{\includegraphics[width=0.75\textwidth]{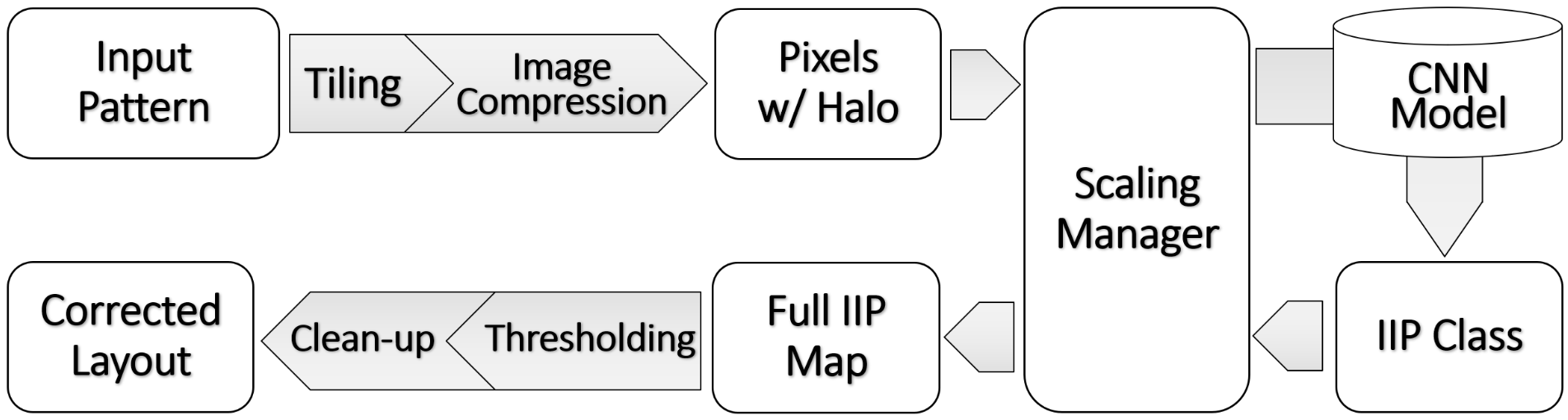}}
	\caption{Flow chart for model deployment in photomask optimization application.}
	\label{eval}
\end{figure*}

Fig. \ref{eval} shows the flow chart for the model deployment and post processing phase. First, the target pattern is tiled into smaller images representing the corrected pixels and their respective ID distance. Such images are then compressed to match the input size of the CNN model. 

The scaling manager organizes the distribution of the pixels data and the predicted IIP class to and from the CNN model instances respectively. The CNN model evaluates the pixel data \(p(x_p,y_p)\) and predicts the IIP class equivalent to it \(IIP(x_p,y_p)\). The calculated IIP values are then arranged in the respective order of their input pixels, which ultimately forms the complete IIP map corresponding to the target pattern.

The IIP map is then converted to a binary pattern by applying a threshold to it, hence defining the output clear-cut delineated geometries. Post-processing is then applied to convert the obtained geometries to standard layout format and apply any required clean-ups to meet restrictions on minimum pattern area or edge length.

It is worth noting that for some applications the IIP map itself can be the desired outputs, such as electrical stress or heat map, while others require further processing to obtain binary output representation such as the corrected photomask in case of RET applications.

\section{Case Study}\label{cs}
For the sake of this paper, we demonstrate the TPM-RET flow by comparing end-to-end results based on OPC and SRAF correction of a 32nm metal test-pattern. The reference tool is  Calibre\texttrademark{} pxOPC\texttrademark{} which is used to generate the ideal photomasks. 

The test patterns are divided into groups according to topology, width and space. The model training flow only uses pixel data from the 40nm and 140nm isolated line and line-space patterns. The patterns from other dimensions and topologies are used only for testing and results comparison.

The TPM-RET flow prototype is implemented using Python 3.9\cite{van1995python} and Calibre\texttrademark{} DesignRev\texttrademark{} internal scripting language. ML model training is done using Tensorflow and Keras, and the model of choice for this case study is mobileNetV3\cite{howardSearchingMobileNetV32019}. The TPM-RET flow settings for the training loop are listed in table \ref{tab2}. 

\begin{table}[htbp]
	\begin{center}
		\caption{TPM-RET Settings. }
		\begin{tabular}{|c|c|}
			\hline
			\textbf{Parameter Name} & \textbf{Value} \\
			\hline
			\# Classes & 100 \\
			\hline
			Pixels/\(nm^2\) & 4 \\
			\hline
			ID  & 400nm \\
			\hline
			IIK  & Optimized 32nm IIK \\
			\hline
		\end{tabular}
		\label{tab2}
	\end{center}
\end{table}

The TPM-RET for data-preparation, IIP map prediction and cleanup was run on a 128 Intel\textregistered{} Xeon\textregistered{} CPU E7-4830 2.1GHz CPU cores and 384GB RAM machine. The ML model training was done on a separate machine with an Nvidia\textregistered{} Quadro\textregistered{} RTX 4000 GPU and 32GB RAM, which is the only step in this prototype that needs to be executed in a GPU-based environment.

\begin{figure*}[hbtp]
	\centering
	\includegraphics[width=0.90\textwidth]{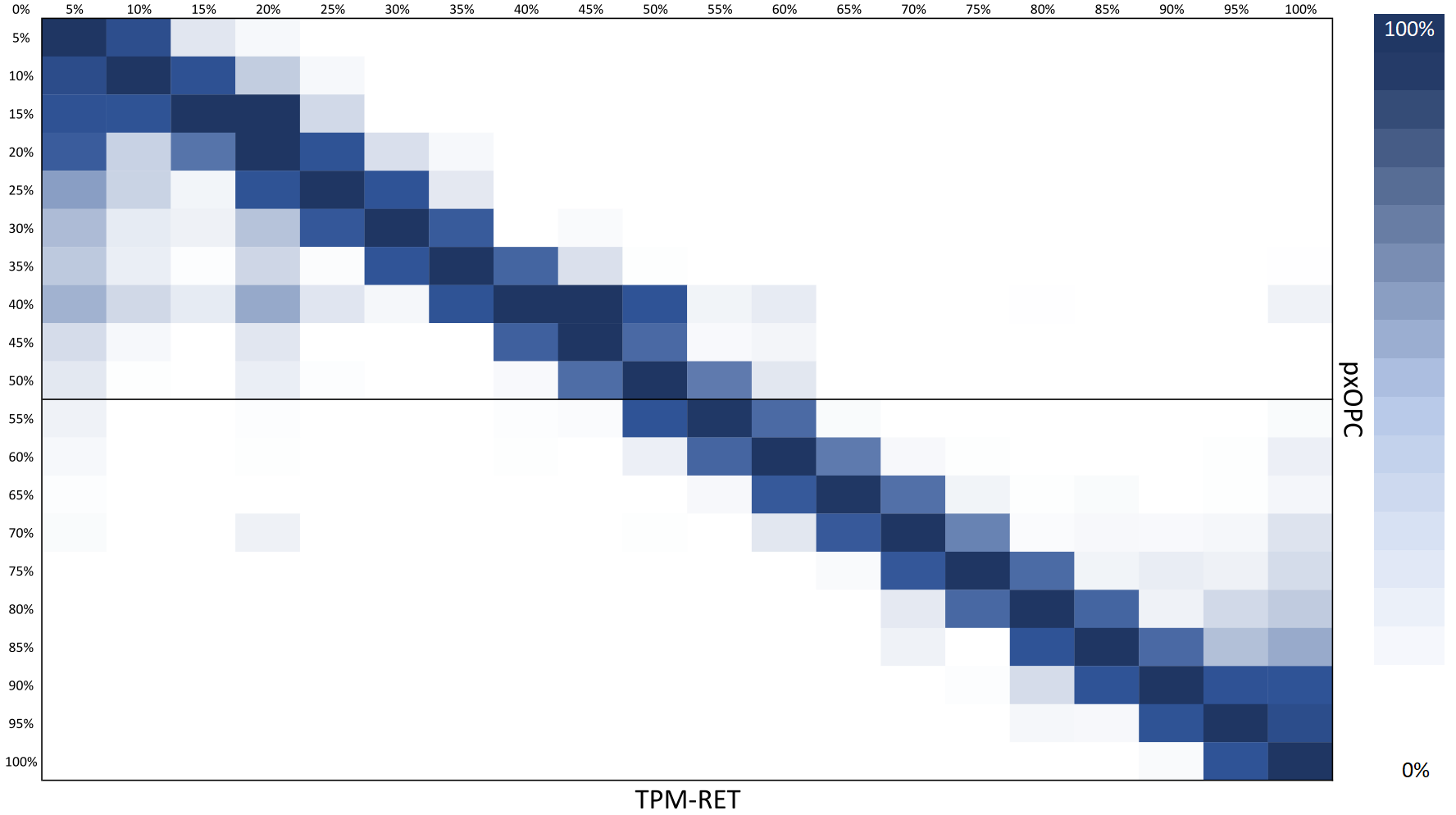}
	\caption{Confusion Matrix for TPM-RET IIP predictions versus the IIP reference generated from pxOPC photomask.}
	\label{cMat}
\end{figure*}

The testing dataset is used to generate a confusion matrix, fig. \ref{cMat}, to compare the accuracy of TPM-RET versus the reference pxOPC\texttrademark{}. The matrix shows well correlation with the reference results and illustrates good ML model stability.

\begin{figure*}[bhtp]
	\centering
	\includegraphics[width=0.90\textwidth]{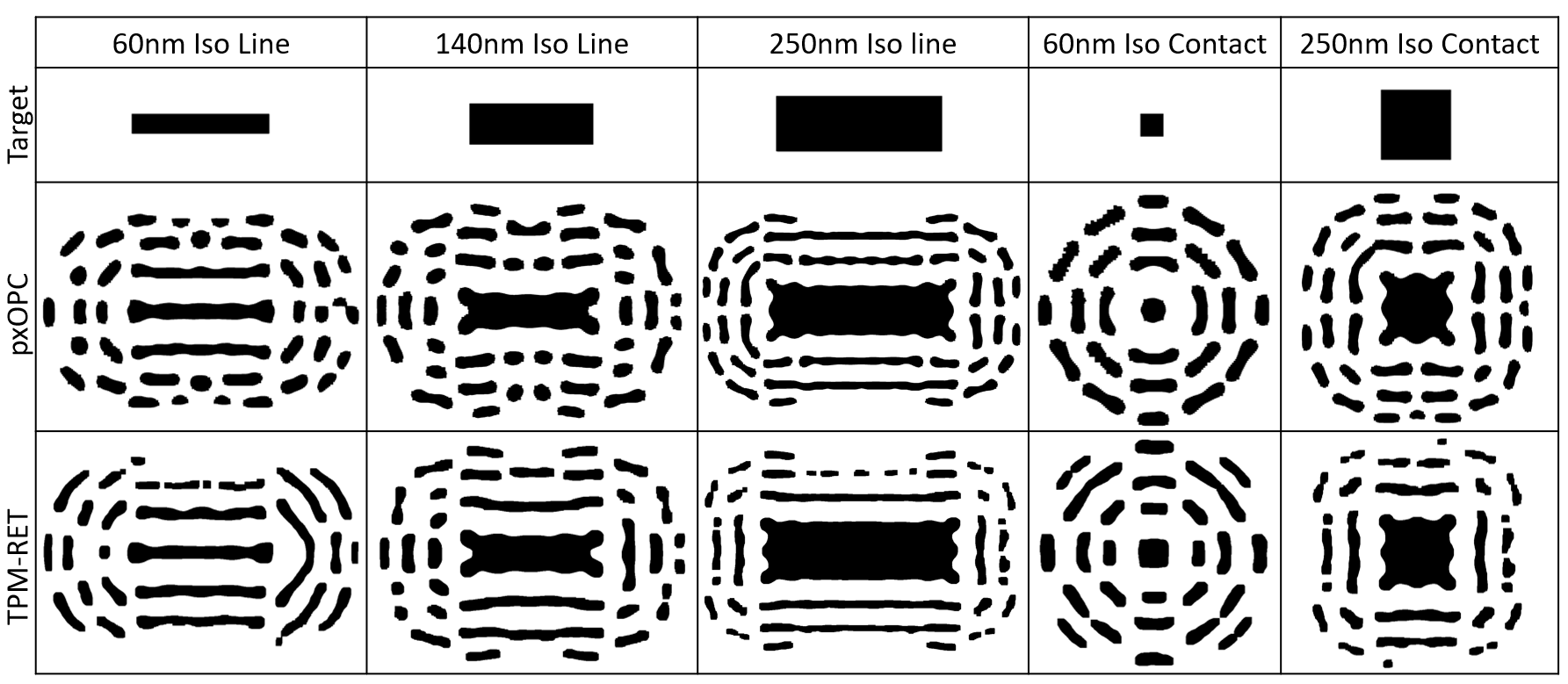}
	\caption{TPM-RET flow results comparison with the reference pxOPC\texttrademark{} for Isolated patterns.}
	\label{results1}
\end{figure*}

\begin{figure*}[hbtp]
	\centering
	\includegraphics[width=0.90\textwidth]{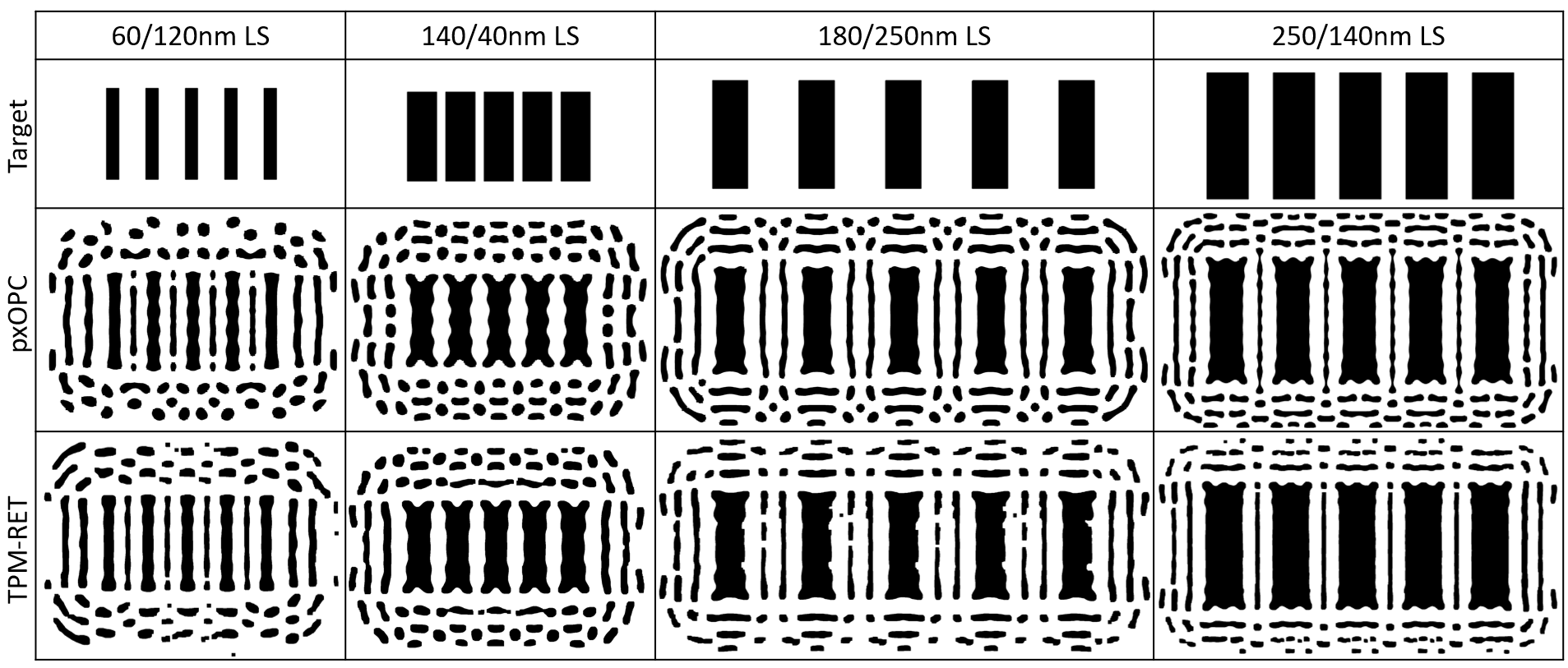}
	\caption{TPM-RET flow results comparison with  pxOPC\texttrademark{} for Line-Space patterns.}
	\label{results2}
\end{figure*}

TPM-RET flow is also used to perform end-to-end correction for the test patterns. Fig. \ref{results1} shows the flow results for isolated lines and squares, while fig. \ref{results2} shows its results for line-space patterns. The results show the correlation between the pxOPC\texttrademark{} and TPM-RET photomasks for various pattern dimensions although the mobileNetV3 model was only trained for \(40nm\) and \(140nm\) line-space and isolated line patterns.

\section{Discussion: TPM-RET Advantages}\label{discuss}
\subsection{Scalability}\label{scal}
Since the model evaluation for a given pixel does not depend on or affect other pixels, we take advantage of that by distributing the execution over multiple processing units. Furthermore, the full-chip computation process can be efficiently distributed over CPU grids, since the TPM-RET flow use computationally sparing ML models. This is a great advantage as silicon foundries owns huge CPU grid systems and it is their preferred execution platform.

\begin{figure*}[hbtp]
	\centerline{\includegraphics[width=0.70\textwidth]{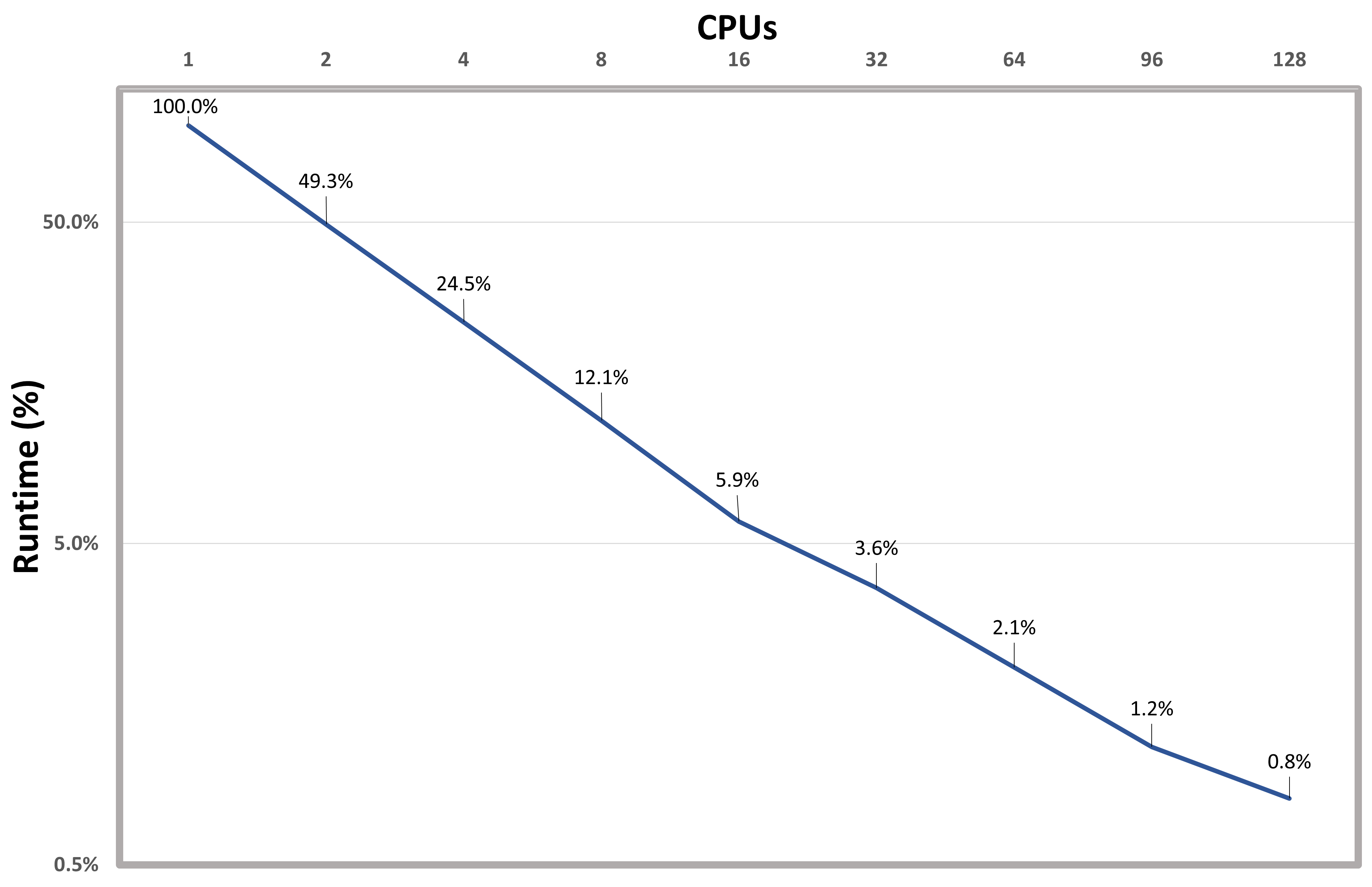}}
	\caption{Logscale chart for TPM-RET runtime scaling versus number of CPUs.}
	\label{scale}
\end{figure*}

TPM-RET flow scales well with great efficiency as shown in the log-scale chart in fig. \ref{scale} as the number of CPUs increase. The chart shows a \(125x\) runtime speed-up running when using 128 CPUs versus a single CPU, providing over \(97\%\) scaling efficiency. We foresee the TPM-RET flow scaling at the same rate for larger number of CPU units. The experiment was done on the same 128 CPU cores machine from section \ref{cs}. 

The distribution process it self is controlled by the scaling manager module, shown in fig  \ref{eval}. This module is implemented using custom python code to manage the slicing and reassembly of chip data with minimal runtime overhead. 

\subsection{Consistency}\label{cons}
TMP-RET flow does not use panning and window slicing to fit the input patterns into its input size. This eliminates the boundary conflicts and the need for stitching patterns spanning across multiple correction windows, hence the flow will not generate inconsistent correction due to chip slicing.

Furthermore, the TPM-RET flow does not generate mismatching halo around identical patterns, since handling the input patterns on a single pixel basis eliminates the case of asymmetrical ID. This guarantees identical halo around identical pixels, which in turn necessitates same inference result.

\subsection{Re-Correction}\label{recor}
When a photomask error is caught in a traditional RET flow, a photomask correction process is performed by either tuning the correction recipe and rerunning the flow unto the erroneous areas or manually if the change scope is manageable. 

TPM-RET allows for such re-correction process by allowing a fine-tuned ML model to run over the pixels representing the erroneous areas. Furthermore, the TPM-RET flow can handle multiple ML models performing correction at different chip areas by defining the execution ranges of operation for each of them.

\subsection{Full-Chip Ready}
TPM-RET flow addresses concerns regarding ML-RET full-chip processing. First, it does not require stitching or window splitting, which eliminates most of the post processing optimizations and also clears the inconsistency associated with them as discussed in \ref{cons}.

The flow is also able to scale over large number of CPUs due to its true-pixel-based nature. This appeals to the CPU infrastructure preferred at silicon foundries, hence eliminating the friction of adopting GPU based infrastructure as a necessity to use ML-based solutions.

\subsection{Recapturing Process Information}
The data available in a reference photomask is binary where as any pixel on such mask can be opaque or bright. Even if the reference is an ideal ILT photomask, the information of the process and the intermediate steps leading to this mask are concealed. 

The IIP algorithm allows the TPM-RET flow to uncover information masked underneath the binary photomask and provides a gradual profile describing transitions between the bright and opaque areas. This gives the model more information to work with and makes it less prone to overfitting.

\subsection{End-to-End Solution}
As discussed in \ref{part}, traditional RET flows splits the mask correction into consequent steps due to historical and flow design reasons. ML-RET applications do not have to follow such footsteps, especially with the complications and potential accuracy setbacks that come with it. 

With that in mind, TPM-RET flow is designed as an end-to-end framework for any lithography applications as long as it initiates from an input layout pattern and ends with a similar sized output.

\subsection{Flexibility}
TPM-RET platform can handle different accuracy and resolution settings with minor or no changes. Prediction accuracy can be controlled by tuning the number of classes, the number of pixels per \(nm^2\), the ID and the IIK. Such changes can be accounted for without any major modifications to the flow.

TPM-RET flow can as well as fit plethora of full chip applications. TPM-RET flow can account for different correction and modeling tasks by replacing the ``ideal photomask'' block in fig. \ref{train} with the desired reference and selecting a proper IIK. For a layout simulator, the reference should be an accurate simulation of the target pattern and a suitable IIK to generate the corresponding IIP map. Similarly, an electrical stress simulator takes a full-chip stress map as its ideal reference which can also be used as the IIP map, and the flow output will not require thresholding or clean-up in that case.

\section{Future Work}\label{future}
Our future work plans for TPM-RET include the following items:
\begin{enumerate}
	\item Convert the TPM-RET Python code to C/C++\cite{ISO:1998:IIP} to further improve execution runtime and scalability.
	\item Develop an algorithm to filter and select the pixels generated from the TPM-RET data-preparation phase.
	\item Implement re-correction and multi-model execution features.
\end{enumerate}

\section{Conclusions}\label{conclusions}
In this study, we reviewed the traditional RET techniques as well as ML-RET state-of-the-art. We highlighted the main obstacles preventing production-ready ML-RET technologies. We then introduced TPM-RET flow, a novel ML flow capable of performing end-to-end lithography correction in a consistent, scalable, and production friendly manner. We demonstrated the flow results based on a 32nm immersion lithography technology by performing an end-to-end correction targeting both OPC and SRAF. Finally, we shared our future work plans for the TPM-RET flow.





\bibliographystyle{model1-num-names}

\bibliography{lib}

\begin{thebibliography}{25}
\expandafter\ifx\csname natexlab\endcsname\relax\def\natexlab#1{#1}\fi
\providecommand{\url}[1]{\texttt{#1}}
\providecommand{\href}[2]{#2}
\providecommand{\path}[1]{#1}
\providecommand{\DOIprefix}{doi:}
\providecommand{\ArXivprefix}{arXiv:}
\providecommand{\URLprefix}{URL: }
\providecommand{\Pubmedprefix}{pmid:}
\providecommand{\doi}[1]{\href{http://dx.doi.org/#1}{\path{#1}}}
\providecommand{\Pubmed}[1]{\href{pmid:#1}{\path{#1}}}
\providecommand{\bibinfo}[2]{#2}
\ifx\xfnm\relax \def\xfnm[#1]{\unskip,\space#1}\fi
\bibitem[{Mack(2007)}]{mackFundamentalPrinciplesOptical2007}
\bibinfo{author}{C.~A. Mack}, \bibinfo{title}{Fundamental Principles of Optical Lithography: The Science of Microfabrication}, \bibinfo{publisher}{{Wiley}}, \bibinfo{address}{{Chichester, West Sussex, England ; Hoboken, NJ, USA}}, \bibinfo{year}{2007}.
\bibitem[{Wong(2001)}]{wongResolutionEnhancementTechniques2001}
\bibinfo{author}{A.~K. Wong}, \bibinfo{title}{Resolution {{Enhancement Techniques}} in {{Optical Lithography}}}, \bibinfo{publisher}{{SPIE}}, \bibinfo{year}{2001}. \DOIprefix\doi{10.1117/3.401208}.
\bibitem[{Lin et~al.(2009)Lin, Donnelly, and Schulze}]{linModelBasedMask2009}
\bibinfo{author}{T.~Lin}, \bibinfo{author}{T.~Donnelly}, \bibinfo{author}{S.~Schulze},
\newblock \bibinfo{title}{Model based mask process correction and verification for advanced process nodes},
\newblock in: \bibinfo{editor}{H.~J. Levinson}, \bibinfo{editor}{M.~V. Dusa} (Eds.), \bibinfo{booktitle}{{{SPIE Advanced Lithography}}}, \bibinfo{address}{{San Jose, California, USA}}, \bibinfo{year}{2009}, p. \bibinfo{pages}{72742A}. \DOIprefix\doi{10.1117/12.814362}.
\bibitem[{Yang et~al.(2018)Yang, Li, Ma, Yu, and Y.~Young}]{yangGANOPCMaskOptimization2018}
\bibinfo{author}{H.~Yang}, \bibinfo{author}{S.~Li}, \bibinfo{author}{Y.~Ma}, \bibinfo{author}{B.~Yu}, \bibinfo{author}{E.~F. Y.~Young},
\newblock \bibinfo{title}{{{GAN-OPC}}: {{Mask Optimization}} with {{Lithography-guided Generative Adversarial Nets}}},
\newblock in: \bibinfo{booktitle}{2018 55th {{ACM}}/{{ESDA}}/{{IEEE Design Automation Conference}} ({{DAC}})}, \bibinfo{publisher}{{IEEE}}, \bibinfo{address}{{San Francisco, CA}}, \bibinfo{year}{2018}, pp. \bibinfo{pages}{1--6}. \DOIprefix\doi{10.1109/DAC.2018.8465816}.
\bibitem[{Chen et~al.(2021)Chen, Yu, Liu, Ma, and Yu}]{chenDevelSetDeepNeural2021}
\bibinfo{author}{G.~Chen}, \bibinfo{author}{Z.~Yu}, \bibinfo{author}{H.~Liu}, \bibinfo{author}{Y.~Ma}, \bibinfo{author}{B.~Yu},
\newblock \bibinfo{title}{{{DevelSet}}: {{Deep Neural Level Set}} for {{Instant Mask Optimization}}},
\newblock in: \bibinfo{booktitle}{2021 {{IEEE}}/{{ACM International Conference On Computer Aided Design}} ({{ICCAD}})}, \bibinfo{publisher}{{IEEE}}, \bibinfo{address}{{Munich, Germany}}, \bibinfo{year}{2021}, pp. \bibinfo{pages}{1--9}. \DOIprefix\doi{10.1109/ICCAD51958.2021.9643464}.
\bibitem[{Jiang et~al.(2022)Jiang, Liu, Ma, Yu, and Young}]{jiangNeuralILTMigratingILT2022}
\bibinfo{author}{B.~Jiang}, \bibinfo{author}{L.~Liu}, \bibinfo{author}{Y.~Ma}, \bibinfo{author}{B.~Yu}, \bibinfo{author}{E.~F.~Y. Young},
\newblock \bibinfo{title}{Neural-{{ILT}} 2.0: {{Migrating ILT}} to {{Domain-Specific}} and {{Multitask-Enabled Neural Network}}},
\newblock \bibinfo{journal}{IEEE Transactions on Computer-Aided Design of Integrated Circuits and Systems} \bibinfo{volume}{41} (\bibinfo{year}{2022}) \bibinfo{pages}{2671--2684}.
\bibitem[{Ciou et~al.(2022)Ciou, Hu, Tsai, Hsuan, Yang, Yang, and Chen}]{ciouMachineLearningOPC2022}
\bibinfo{author}{W.~Ciou}, \bibinfo{author}{T.~Hu}, \bibinfo{author}{Y.-Y. Tsai}, \bibinfo{author}{C.-T. Hsuan}, \bibinfo{author}{E.~Yang}, \bibinfo{author}{T.-H. Yang}, \bibinfo{author}{K.-C. Chen},
\newblock \bibinfo{title}{Machine learning {{OPC}} with generative adversarial networks},
\newblock in: \bibinfo{editor}{N.~V. Lafferty}, \bibinfo{editor}{R.-H. Kim} (Eds.), \bibinfo{booktitle}{{{DTCO}} and {{Computational Patterning}}}, \bibinfo{publisher}{{SPIE}}, \bibinfo{address}{{San Jose, United States}}, \bibinfo{year}{2022}, p.~\bibinfo{pages}{27}. \DOIprefix\doi{10.1117/12.2606715}.
\bibitem[{Chen et~al.(2022)Chen, Chen, Sun, Ma, Yang, and Yu}]{chenDAMODeepAgile2022}
\bibinfo{author}{G.~Chen}, \bibinfo{author}{W.~Chen}, \bibinfo{author}{Q.~Sun}, \bibinfo{author}{Y.~Ma}, \bibinfo{author}{H.~Yang}, \bibinfo{author}{B.~Yu},
\newblock \bibinfo{title}{{{DAMO}}: {{Deep Agile Mask Optimization}} for {{Full-Chip Scale}}},
\newblock \bibinfo{journal}{IEEE Transactions on Computer-Aided Design of Integrated Circuits and Systems} \bibinfo{volume}{41} (\bibinfo{year}{2022}) \bibinfo{pages}{3118--3131}.
\bibitem[{Alawieh et~al.(2021)Alawieh, Lin, Zhang, Li, Huang, and Pan}]{alawiehGANSRAFSubresolutionAssist2021}
\bibinfo{author}{M.~B. Alawieh}, \bibinfo{author}{Y.~Lin}, \bibinfo{author}{Z.~Zhang}, \bibinfo{author}{M.~Li}, \bibinfo{author}{Q.~Huang}, \bibinfo{author}{D.~Z. Pan},
\newblock \bibinfo{title}{{{GAN-SRAF}}: {{Subresolution Assist Feature Generation Using Generative Adversarial Networks}}},
\newblock \bibinfo{journal}{IEEE Transactions on Computer-Aided Design of Integrated Circuits and Systems} \bibinfo{volume}{40} (\bibinfo{year}{2021}) \bibinfo{pages}{373--385}.
\bibitem[{Goodfellow et~al.(2014)Goodfellow, {Pouget-Abadie}, Mirza, Xu, {Warde-Farley}, Ozair, Courville, and Bengio}]{goodfellowGenerativeAdversarialNetworks2014}
\bibinfo{author}{I.~J. Goodfellow}, \bibinfo{author}{J.~{Pouget-Abadie}}, \bibinfo{author}{M.~Mirza}, \bibinfo{author}{B.~Xu}, \bibinfo{author}{D.~{Warde-Farley}}, \bibinfo{author}{S.~Ozair}, \bibinfo{author}{A.~Courville}, \bibinfo{author}{Y.~Bengio}, \bibinfo{title}{Generative {{Adversarial Networks}}}, \bibinfo{year}{2014}. \DOIprefix\doi{10.48550/ARXIV.1406.2661}.
\bibitem[{Ciou et~al.(2021)Ciou, Hu, Tsai, Hsuan, Yang, Yang, and Chen}]{ciouSRAFPlacementGenerative2021}
\bibinfo{author}{W.~Ciou}, \bibinfo{author}{T.~Hu}, \bibinfo{author}{Y.-Y. Tsai}, \bibinfo{author}{T.~Hsuan}, \bibinfo{author}{E.~Yang}, \bibinfo{author}{T.~Yang}, \bibinfo{author}{K.~Chen},
\newblock \bibinfo{title}{{{SRAF}} placement with generative adversarial network},
\newblock in: \bibinfo{editor}{S.~Owa}, \bibinfo{editor}{M.~C. Phillips} (Eds.), \bibinfo{booktitle}{Optical {{Microlithography XXXIV}}}, \bibinfo{publisher}{{SPIE}}, \bibinfo{address}{{Online Only, United States}}, \bibinfo{year}{2021}, p.~\bibinfo{pages}{3}. \DOIprefix\doi{10.1117/12.2581334}.
\bibitem[{Matsunawa et~al.(2016)Matsunawa, Yu, and Pan}]{matsunawaOpticalProximityCorrection2016}
\bibinfo{author}{T.~Matsunawa}, \bibinfo{author}{B.~Yu}, \bibinfo{author}{D.~Z. Pan},
\newblock \bibinfo{title}{Optical proximity correction with hierarchical {{Bayes}} model},
\newblock \bibinfo{journal}{Journal of Micro/Nanolithography, MEMS, and MOEMS} \bibinfo{volume}{15} (\bibinfo{year}{2016}) \bibinfo{pages}{021009}.
\bibitem[{Kwon et~al.(2019)Kwon, Song, and Shin}]{kwonOpticalProximityCorrection2019}
\bibinfo{author}{Y.~Kwon}, \bibinfo{author}{Y.~Song}, \bibinfo{author}{Y.~Shin},
\newblock \bibinfo{title}{Optical proximity correction using bidirectional recurrent neural network ({{BRNN}})},
\newblock in: \bibinfo{editor}{J.~P. Cain}, \bibinfo{editor}{C.-M. Yuan} (Eds.), \bibinfo{booktitle}{Design-{{Process-Technology Co-optimization}} for {{Manufacturability XIII}}}, \bibinfo{publisher}{{SPIE}}, \bibinfo{address}{{San Jose, United States}}, \bibinfo{year}{2019}, p.~\bibinfo{pages}{12}. \DOIprefix\doi{10.1117/12.2515159}.
\bibitem[{Geng et~al.(2020)Geng, Zhong, Yang, Ma, Mitra, and Yu}]{gengSRAFInsertionSupervised2020}
\bibinfo{author}{H.~Geng}, \bibinfo{author}{W.~Zhong}, \bibinfo{author}{H.~Yang}, \bibinfo{author}{Y.~Ma}, \bibinfo{author}{J.~Mitra}, \bibinfo{author}{B.~Yu},
\newblock \bibinfo{title}{{{SRAF Insertion}} via {{Supervised Dictionary Learning}}},
\newblock \bibinfo{journal}{IEEE Transactions on Computer-Aided Design of Integrated Circuits and Systems} \bibinfo{volume}{39} (\bibinfo{year}{2020}) \bibinfo{pages}{2849--2859}.
\bibitem[{Shim and Shin(2016)}]{shimEtchProximityCorrection2016}
\bibinfo{author}{S.~Shim}, \bibinfo{author}{Y.~Shin},
\newblock \bibinfo{title}{Etch proximity correction through machine-learning-driven etch bias model},
\newblock in: \bibinfo{editor}{Q.~Lin}, \bibinfo{editor}{S.~U. Engelmann} (Eds.), \bibinfo{booktitle}{{{SPIE Advanced Lithography}}}, \bibinfo{address}{{San Jose, California, United States}}, \bibinfo{year}{2016}, p. \bibinfo{pages}{97820O}. \DOIprefix\doi{10.1117/12.2219057}.
\bibitem[{Meng et~al.(2021)Meng, Kim, Guo, Shu, Zhang, and Liu}]{mengMachineLearningModels2021}
\bibinfo{author}{Y.~Meng}, \bibinfo{author}{Y.-C. Kim}, \bibinfo{author}{S.~Guo}, \bibinfo{author}{Z.~Shu}, \bibinfo{author}{Y.~Zhang}, \bibinfo{author}{Q.~Liu},
\newblock \bibinfo{title}{Machine {{Learning Models}} for {{Edge Placement Error Based Etch Bias}}},
\newblock \bibinfo{journal}{IEEE Transactions on Semiconductor Manufacturing} \bibinfo{volume}{34} (\bibinfo{year}{2021}) \bibinfo{pages}{42--48}.
\bibitem[{Sharma et~al.(2019)Sharma, Durvasula, Rao, Bork, Sharma, Mishra, and Buck}]{sharmaMachineLearningGuided2019}
\bibinfo{author}{M.~Sharma}, \bibinfo{author}{B.~S. Durvasula}, \bibinfo{author}{N.~Rao}, \bibinfo{author}{I.~Bork}, \bibinfo{author}{R.~Sharma}, \bibinfo{author}{K.~Mishra}, \bibinfo{author}{P.~Buck},
\newblock \bibinfo{title}{Machine learning guided curvilinear {{MPC}}},
\newblock in: \bibinfo{editor}{J.~H. Rankin}, \bibinfo{editor}{M.~E. Preil} (Eds.), \bibinfo{booktitle}{Photomask {{Technology}} 2019}, \bibinfo{publisher}{{SPIE}}, \bibinfo{address}{{Monterey, United States}}, \bibinfo{year}{2019}, p.~\bibinfo{pages}{24}. \DOIprefix\doi{10.1117/12.2538646}.
\bibitem[{Pang(2021)}]{pangInverseLithographyTechnology2021}
\bibinfo{author}{L.~L. Pang},
\newblock \bibinfo{title}{Inverse lithography technology: 30 years from concept to practical, full-chip reality},
\newblock \bibinfo{journal}{Journal of Micro/Nanopatterning, Materials, and Metrology} \bibinfo{volume}{20} (\bibinfo{year}{2021}).
\bibitem[{Ma et~al.(2014)Ma, Wu, Song, Jiang, and Li}]{maFastPixelbasedOptical2014}
\bibinfo{author}{X.~Ma}, \bibinfo{author}{B.~Wu}, \bibinfo{author}{Z.~Song}, \bibinfo{author}{S.~Jiang}, \bibinfo{author}{Y.~Li},
\newblock \bibinfo{title}{Fast pixel-based optical proximity correction based on nonparametric kernel regression},
\newblock \bibinfo{journal}{Journal of Micro/Nanolithography, MEMS, and MOEMS} \bibinfo{volume}{13} (\bibinfo{year}{2014}) \bibinfo{pages}{043007}.
\bibitem[{{Siemens EDA}(2023)}]{siemensedaCalibreComputationalLithography2023}
\bibinfo{author}{{Siemens EDA}}, \bibinfo{title}{Calibre {{Computational Lithography}}}, \bibinfo{howpublished}{https://eda.sw.siemens.com/en-US/ic/calibre-manufacturing/computational-lithography/}, \bibinfo{year}{2023}.
\bibitem[{Abadi et~al.(2015)Abadi, Agarwal, Barham, Brevdo, Chen, Citro, Corrado, Davis, Dean, Devin, Ghemawat, Goodfellow, Harp, Irving, Isard, Jia, Jozefowicz, Kaiser, Kudlur, Levenberg, Man{\'e}, Monga, Moore, Murray, Olah, Schuster, Shlens, Steiner, Sutskever, Talwar, Tucker, Vanhoucke, Vasudevan, Vi{\'e}gas, Vinyals, Warden, Wattenberg, Wicke, Yu, and Zheng}]{tensorflow2015}
\bibinfo{author}{M.~Abadi}, \bibinfo{author}{A.~Agarwal}, \bibinfo{author}{P.~Barham}, \bibinfo{author}{E.~Brevdo}, \bibinfo{author}{Z.~Chen}, \bibinfo{author}{C.~Citro}, \bibinfo{author}{G.~S. Corrado}, \bibinfo{author}{A.~Davis}, \bibinfo{author}{J.~Dean}, \bibinfo{author}{M.~Devin}, \bibinfo{author}{S.~Ghemawat}, \bibinfo{author}{I.~Goodfellow}, \bibinfo{author}{A.~Harp}, \bibinfo{author}{G.~Irving}, \bibinfo{author}{M.~Isard}, \bibinfo{author}{Y.~Jia}, \bibinfo{author}{R.~Jozefowicz}, \bibinfo{author}{L.~Kaiser}, \bibinfo{author}{M.~Kudlur}, \bibinfo{author}{J.~Levenberg}, \bibinfo{author}{D.~Man{\'e}}, \bibinfo{author}{R.~Monga}, \bibinfo{author}{S.~Moore}, \bibinfo{author}{D.~Murray}, \bibinfo{author}{C.~Olah}, \bibinfo{author}{M.~Schuster}, \bibinfo{author}{J.~Shlens}, \bibinfo{author}{B.~Steiner}, \bibinfo{author}{I.~Sutskever}, \bibinfo{author}{K.~Talwar}, \bibinfo{author}{P.~Tucker}, \bibinfo{author}{V.~Vanhoucke}, \bibinfo{author}{V.~Vasudevan}, \bibinfo{author}{F.~Vi{\'e}gas},
  \bibinfo{author}{O.~Vinyals}, \bibinfo{author}{P.~Warden}, \bibinfo{author}{M.~Wattenberg}, \bibinfo{author}{M.~Wicke}, \bibinfo{author}{Y.~Yu}, \bibinfo{author}{X.~Zheng}, \bibinfo{title}{{{TensorFlow}}: {{Large-scale}} machine learning on heterogeneous systems}, \bibinfo{year}{2015}.
\bibitem[{Chollet et~al.(2015)}]{chollet2015keras}
\bibinfo{author}{F.~Chollet}, et~al., \bibinfo{title}{Keras}, \bibinfo{year}{2015}.
\bibitem[{Van~Rossum and Drake~Jr(1995)}]{van1995python}
\bibinfo{author}{G.~Van~Rossum}, \bibinfo{author}{F.~L. Drake~Jr}, \bibinfo{title}{Python Reference Manual}, \bibinfo{publisher}{{Centrum voor Wiskunde en Informatica Amsterdam}}, \bibinfo{year}{1995}.
\bibitem[{Howard et~al.(2019)Howard, Sandler, Chu, Chen, Chen, Tan, Wang, Zhu, Pang, Vasudevan, Le, and Adam}]{howardSearchingMobileNetV32019}
\bibinfo{author}{A.~Howard}, \bibinfo{author}{M.~Sandler}, \bibinfo{author}{G.~Chu}, \bibinfo{author}{L.-C. Chen}, \bibinfo{author}{B.~Chen}, \bibinfo{author}{M.~Tan}, \bibinfo{author}{W.~Wang}, \bibinfo{author}{Y.~Zhu}, \bibinfo{author}{R.~Pang}, \bibinfo{author}{V.~Vasudevan}, \bibinfo{author}{Q.~V. Le}, \bibinfo{author}{H.~Adam}, \bibinfo{title}{Searching for {{MobileNetV3}}}, \bibinfo{year}{2019}. \href{http://arxiv.org/abs/1905.02244}{\tt arXiv:1905.02244}.
\bibitem[{{ISO}(1998)}]{ISO:1998:IIP}
\bibinfo{author}{{ISO}}, \bibinfo{title}{{{ISO}}{\slash}{{IEC}} 14882:1998: {{Programming}} Languages {\textemdash} {{C}}++}, \bibinfo{publisher}{{pub-ISO}}, \bibinfo{address}{{pub-ISO:adr}}, \bibinfo{year}{1998}.

\end{thebibliography}



\end{document}